\documentclass{article}
\usepackage{spconf,amsmath,graphicx}
\usepackage{booktabs}

\title{Speech Emotion Diarization: Which Emotion Appears When?}
\name{Yingzhi Wang$^1$, Mirco Ravanelli$^{2,3,4}$, Alya Yacoubi$^1$\thanks{This research is funded by Zaion.}}
\address{
  $^1$Zaion Lab, Zaion, France\\
  $^2$Mila - Quebec AI Institute, Canada \\
  $^3$Université de Montréal, Canada \\
  $^4$Concordia University, Canada \\
 }
\begin{document}
\ninept
\maketitle
\begin{abstract}
Speech Emotion Recognition (SER) typically relies on utterance-level solutions. However, emotions conveyed through speech should be considered as discrete speech events with definite temporal boundaries, rather than attributes of the entire utterance.
To reflect the fine-grained nature of speech emotions and to unify various fine-grained methods under a single objective, we propose a new task: Speech Emotion Diarization (SED). Just as Speaker Diarization answers the question of ``\textit{Who speaks when?}", Speech Emotion Diarization answers the question of ``\textit{Which emotion appears when?}". To facilitate the evaluation of the performance and establish a common benchmark, we introduce the Zaion Emotion Dataset (ZED), an openly accessible speech emotion dataset that includes non-acted emotions recorded in real-life conditions, along with manually annotated boundaries of emotion segments within the utterance.
%for each speech recording. %Emotion Diarization Error Rate (EDER) is proposed for evaluation. %We have also provided performant baselines and open-sourced the code and the pre-trained models which allow the extraction of fine-grained emotion embeddings.
We provide competitive baselines and open-source the code and the pre-trained models. 
%for the extraction of finely-tuned emotion embeddings.
\end{abstract}
\begin{keywords}
speech emotion diarization, emotion recognition, Zaion Emotion Dataset, emotion diarization error rate
\end{keywords}
\section{Introduction}
\label{sec:intro}

Speech emotion recognition has consistently been regarded as a complex task, primarily owing to two factors: the subjective, complex, and temporally fine-grained expression of speech emotion and the difficulty in finding effective feature representations \cite{survey, review1, review2, review3}. In the past few years, the progress in deep learning has contributed to notable improvements in the performance of emotion recognition systems by leveraging highly effective features extracted from deep neural networks \cite{capsule, benchmark, spk-norm}. Rather than exploring the emotional feature representations, the main objective of this paper is to investigate the fine time granularity of speech emotions.
%However, regarding the fine-grained nature of speech emotion, most of the solutions are still at utterance-level. Some works performed frame-level emotion recognition \cite{dimensional1, dimensional2} and proposed frame-level datasets \cite{SEWA, RECOLA, SEMAINE}, but most of these works were based on dimensional modeling \cite{dim1, dim2, pad}. 

According to research in psychology, two of the most popular emotion modeling approaches can be distinguished: the categorical approach and the dimensional approach \cite{survey}. The dimensional approach \cite{dim1, dim2, pad} has increasingly gained widespread interest and attention because it can better reflect the complexity of some non-basic, subtle, and rather complex emotions like thinking and embarrassment. Frame-level solutions \cite{dimensional1, dimensional2} and datasets \cite{SEWA, RECOLA, SEMAINE} based on the dimensional approach have been proposed to study the temporal interdependence of emotion states. Compared with the dimensional approach, the categorical approach \cite{categorical} has been the most commonly adopted approach thanks to its universality, intuitiveness, and facility for annotation \cite{review3}. However, in terms of temporal granularity, the majority of the solutions still target utterance-level tasks \cite{review1, review3} and are evaluated on utterance-level datasets and with utterance-level metrics, despite the application of frame-level methodologies.
% However, compared with the categorical modeling approach, the dimensional modeling approach is difficult to be widely applied in the industry due to the lack of intuitiveness and high annotation difficulty. %In this paper, we focus on the categorical modeling approaches as little research have been done so far in this field and there is a lack of common benchmarks and dataset for this task of practical interest.
%Some previous studies have been done on categorical modeling approaches for emotion recognition. Some works
%have indeed noticed the fine-grained nature of speech emotions and have proposed some fine-grained emotion recognition methods \cite{emotion-ctc, local-attention, attention-pooling}. However, previous methods are still applied to utterance-level tasks and were evaluated on utterance-level metrics and datasets. Therefore, the effectiveness and innovation of these studies cannot be truly reflected. 

\begin{figure}[t!]
  \centering
  \includegraphics[width=\linewidth]{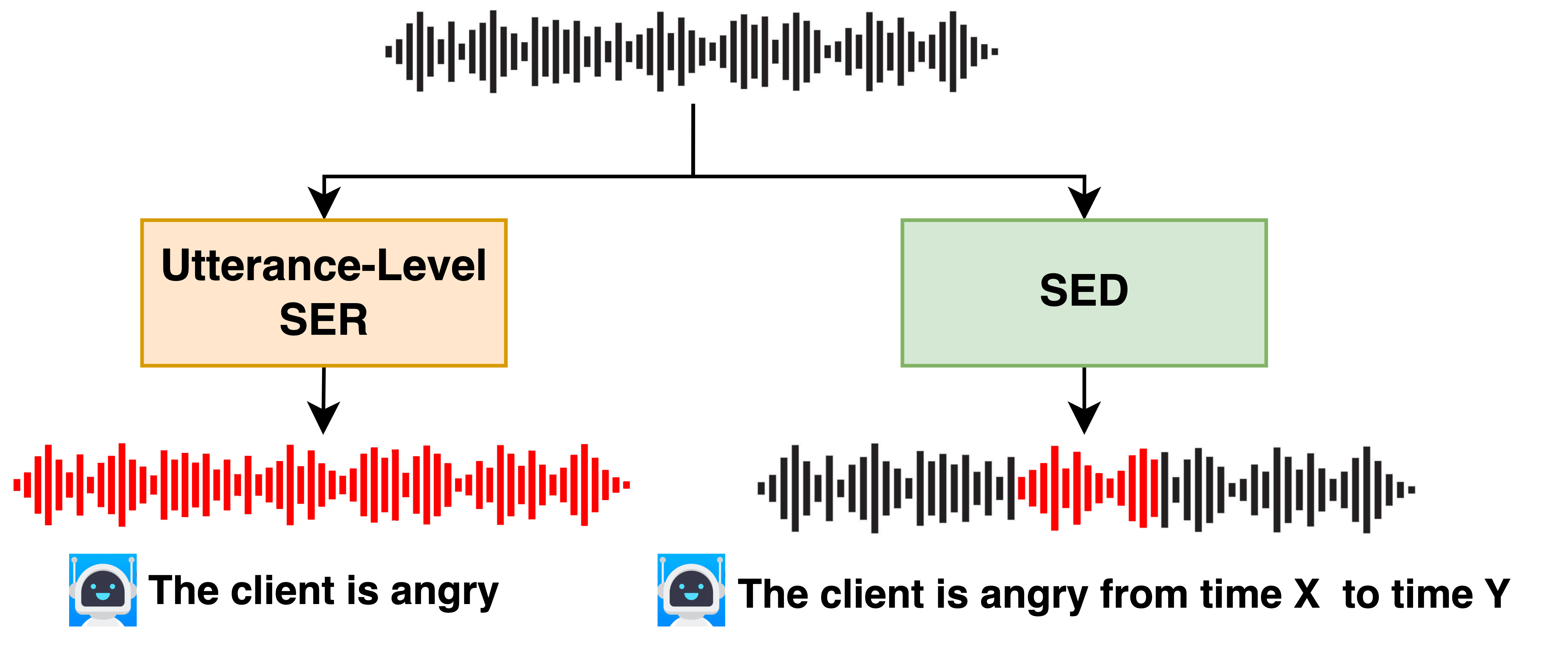}
  \caption{Different goals for utterance-level Speech Emotion Recognition (SER) and Speech Emotion Diarization (SED).  While the former focuses on recognizing the emotions conveyed in an entire utterance, the latter goes beyond by not only identifying the emotions but also accurately locating their boundaries.}
  \label{fig:ser_sed}
\end{figure}

% Mirco's version
This paper focuses on the categorical modeling approach for fine-grained emotion recognition, an area that has received little research attention despite its long-recognized importance \cite{emotion-ctc, video}. 
%For categorical modeling approaches, although some works have noticed the fine-grained nature of speech emotions and have proposed some fine-grained emotion recognition methods \cite{emotion-ctc, local-attention, attention-pooling},  these methods are still applied to utterance-level tasks and were evaluated on utterance-level metrics and datasets. Therefore, the effectiveness and innovation of these studies cannot be truly reflected.
% Mirco's version
Our work aims to address a major issue that has hindered extensive research into fine-grained emotion recognition. Specifically, we identified a lack of common benchmarks, with well-defined tasks, evaluation datasets and metrics. To address this gap, we propose a Speech Emotion Diarization (SED) task.
While standard speech emotion recognition focuses on identifying the emotion that corresponds to a given utterance, the proposed Speech Emotion Diarization task aims to simultaneously identify the correct emotions and their corresponding boundaries. This task is particularly important for capturing the fine-grained nature of speech emotions and for providing a better causality of the emotion predictions. In the context of industrial applications, the Speech Emotion Diarization task presents an opportunity to gain greater precision in emotion analysis and to track the evolution of emotion over time. A comparison between standard utterance-level Speech Emotion Recognition and the proposed Speech Emotion Diarization can be found in Figure \ref{fig:ser_sed}. Inspired by Diarization Error Rate (DER) which is commonly used in Speaker Diarization, we have defined the Emotion Diarization Error Rate (EDER) for the evaluation of the proposed SED task. We also release the Zaion Emotion Dataset (ZED), a fine-grained emotion dataset that is openly accessible to the research community. The ZED dataset includes discrete categorical labels for emotional segments within each utterance, as well as manually annotated boundaries for each emotion segment.

In summary, our work brings the following contributions to the field of fine-grained speech emotion recognition:

\begin{itemize}

\item We propose a Speech Emotion Diarization task that provides a clear and well-defined benchmark to unify various frame-level methods under a diarization objective.

\item We introduce a novel evaluation metric, EDER, that takes into account both the quality of the emotion classification and the accuracy of the detection of emotion boundaries.

\item We release the Zaion Emotion Dataset\footnote{https://zaion.ai/en/resources/zaion-lab-blog/zaion-emotion-dataset/}, a high-quality, manually-annotated, fine-grained emotion dataset that includes non-acted/in-the-wild emotions. The dataset is freely available to the research community, providing a common benchmark for evaluation and encouraging further research in this field.

\item We open-source the code and the pre-trained models \footnote{https://github.com/speechbrain/speechbrain/tree/develop/recipes/} on the popular SpeechBrain toolkit \cite{SpeechBrain}.

\end{itemize}

\section{Related Work}
\label{sec:related}

The concept of frame-level processing and dynamic modeling has a long-standing presence in the field of speech emotion recognition.

At an early stage, \cite{framevsturn} investigated dynamic modeling at frame-level for speech emotion recognition, aiming to capture the important information on temporal sub-turn-layers. The frame-level information was classified by a Gaussian Mixture Model to output a frame-wise score. The output scores were added to a super-vector combined with static acoustic features and passed to an SVM for the final utterance-level classification task.

Then, driven by the advancements in deep learning, a widespread incorporation of deep neural networks has been witnessed in frame-level emotion recognition approaches. In \cite{extreme} the authors calculated low-level features for frames and transformed this sequence of features to the sequence of probability distributions over the target emotion labels through a densely connected neural network. Then these probabilities were aggregated into utterance-level features using simple statistics for a final classification.
\cite{realtime} proposed a real-time SER system based on end-to-end deep learning, where a Deep Neural Network (DNN) was used to recognize emotions from a one-second frame of raw speech spectrogram. The system was evaluated at utterance-level by fusing frame-level predictions.
Both \cite{local-attention} and \cite{attention-pooling} noticed the fact that different frames may have different contributions to the overall utterance-level emotion. They proposed attentional pooling strategies on top of DNN frame-level embeddings in order to focus on specific regions of a speech signal that are more emotionally salient. The speech frames were assigned different weights from the attention mechanism based on how emotional they were decided to be. The utilization of frame-wise weights in these works reflects the fact that researchers have begun to put more emphasis on exploring the fine granularity within speech emotions.

Furthermore, some works showed an emerging trend suggesting that a single emotion label is no longer sufficient to adequately describe the emotion complexity of a spoken utterance. \cite{evaluating} introduced a frame-based formulation of Speech Emotion Recognition to model intra-utterance dynamics. Instead of using only the utterance-level emotion class, the researchers added silence into the output classes since silence and unvoiced speech were not removed from the input speech. The approach was evaluated with utterance-level accuracies. During inference, the intra-utterance emotion transitions were demonstrated with the posterior class probabilities.
\cite{emotion-ctc} took into account the problem that even the emotional utterance might contain non-emotional parts. The study designed intra-utterance emotion transition sequences and innovatively used a CTC method to predict the sequence of emotions within one spoken utterance. The emotion sequences were finally collapsed into one emotion label to meet the utterance-level evaluation.

As can be seen from the aforementioned literature, researchers have consistently applied frame-level methodologies in Speech Emotion Recognition and have acknowledged the significance of intra-utterance dynamics. However, a common constraint is that almost all the mainstream categorical datasets are annotated at utterance-level \cite{IEMOCAP, RAVDESS, EMO-DB}. Besides, the absence of a unified fine-grained metric for assessing frame-level systems compounds this limitation. Consequently, the majority of the frame-level approaches had to aggregate the frame-level embeddings or predictions into utterance-level for performance evaluation. This practice results in a loss of the fine-grained nature of speech emotions that frame-level methods could potentially capture.

To address these restrictions, we introduce a fine-grained emotion recognition task together with a fine-grained emotion dataset. In addition, we propose a general metric that allows the evaluation of various frame-level approaches despite different frame lengths utilized.
Through our work, we seek to accomplish two principal objectives:

\begin{itemize}

\item Reveal the fine temporal granularity brought by frame-level approaches by directly assessing them on a fine-grained task along with a fine-grained dataset and a fine-grained evaluation metric.

\item At the same time, establish a common benchmark to compare the intra-utterance dynamic modeling capacities of different frame-level methods.

\end{itemize}

The proposed Speech Emotion Diarization task can be considered as a sub-task of the general Speech Emotion Recognition task, which is specially designed to more effectively accommodate frame-level processing/dynamic modeling methodologies.

% We hope that the power of the frame-level approaches can be directly reflected through a fine-grained task along with a fine-grained dataset and a fine-grained metric.

% Our objective is to enable the direct assessment of the effectiveness of frame-level techniques within a dedicated frame-level dataset and task, eliminating the necessity for aggregation to the utterance level.

\section{Speech Emotion Diarization}
\label{sec:Speech Emotion Diarization}

The Speech Emotion Diarization task takes a spoken utterance as input and aims at identifying the presence of a pre-defined set of emotion candidates, while also determining the time intervals in which they appear.
%The Speech Emotion Diarization task takes an utterance as input and aims to find out if particular emotions are present within the utterance and to determine the time spans in which they appear. 
%Unlike Speaker Diarization which does not require knowledge of the specific identity of each speaker, Speech Emotion Diarization requires identifying and localizing a pre-defined set of emotion candidates. Further, we propose a pipeline for Speech Emotion Diarization in 2.1 and Emotion Diarization Error Rate as an evaluation metric in 2.2.
In the following, we describe the speech emotion diarization pipeline and the proposed evaluation metric.

\subsection{Speech Emotion Diarization Pipeline}

\begin{figure}[t!]
  \centering
  \includegraphics[width=\linewidth]{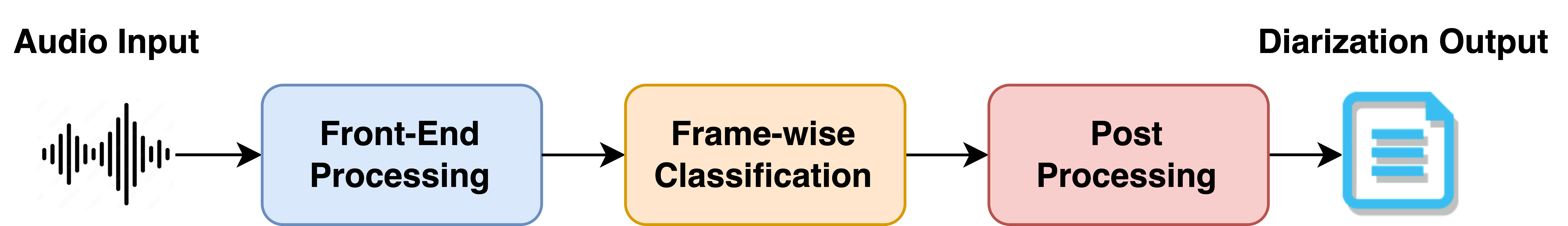}
  \caption{Pipeline for speech emotion diarization system.}
  \label{fig:pipeline}
\end{figure}

% Mirco's version

\begin{figure}[t!]
  \centering
  \includegraphics[width=\linewidth]{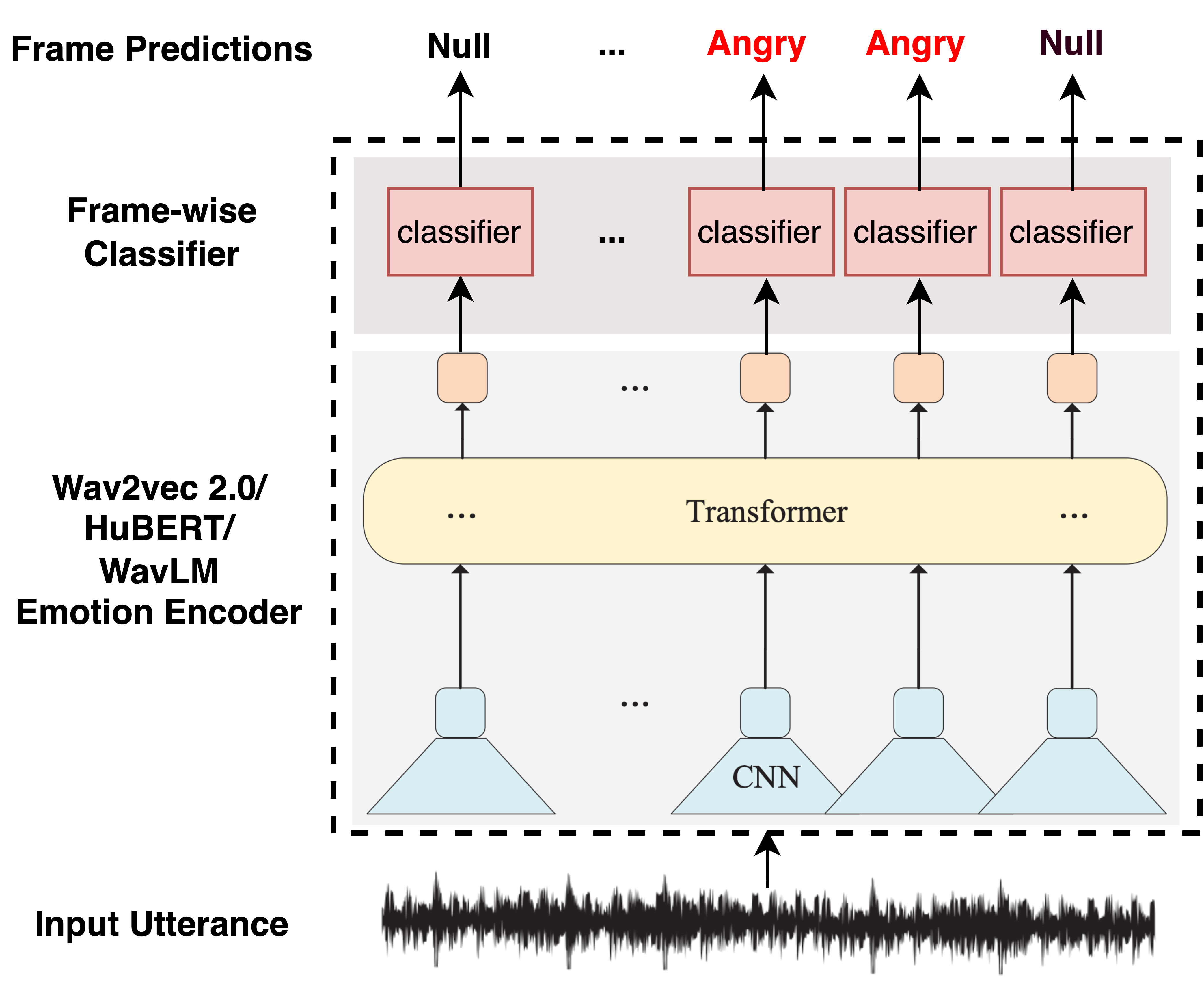}
  \caption{The adopted frame-wise classification model is composed of a fine-grained emotional encoder and a frame-wise classifier.}
  \label{fig:model}
\end{figure}

Figure \ref{fig:pipeline} illustrates the proposed pipeline for Speech Emotion Diarization, which consists of three primary modules: front-end processing, frame-wise classification, and post-processing.
The front-end processing module incorporates pre-processing techniques such as speech enhancement, dereverberation, source separation, and speech segmentation, which are utilized to improve the quality of the speech recordings and determine the appropriate input lengths before feeding the utterances into the frame-wise classification module. The frame-wise classifier predicts the emotional content on a frame-by-frame basis, allowing for a finer time granularity as opposed to classifying an entire input utterance.
A model for frame-wise emotion classification is proposed in Figure \ref{fig:model}, which employs an emotional encoder followed by a linear classifier.
A challenge for fine-grained speech emotion recognition is that different emotional states can vary significantly in duration, with some lasting only a few frames while others could persist for longer periods. To accommodate variable-sized contexts, we adopted a standard self-attention mechanism \cite{attention} by integrating transformer blocks within the emotional encoder. Specifically, we leveraged modern self-supervised models (e.g., Wav2vec 2.0 \cite{wav2vec2}, Hubert \cite{HuBERT}, and WavLM \cite{WavLM}), which have demonstrated superior performance across various tasks including speech emotion recognition \cite{benchmark, wavlm-emo}.

Finally, post-processing techniques can be used to improve the performance of the prediction. For example, the predictions can be filtered based on prior knowledge (e.g., number/position of emotion appearances), or a threshold can be applied to mask low-confidence predictions.

\subsection{Evaluation Metric}
\begin{figure}[t!]
  \centering
  \includegraphics[width=\linewidth]{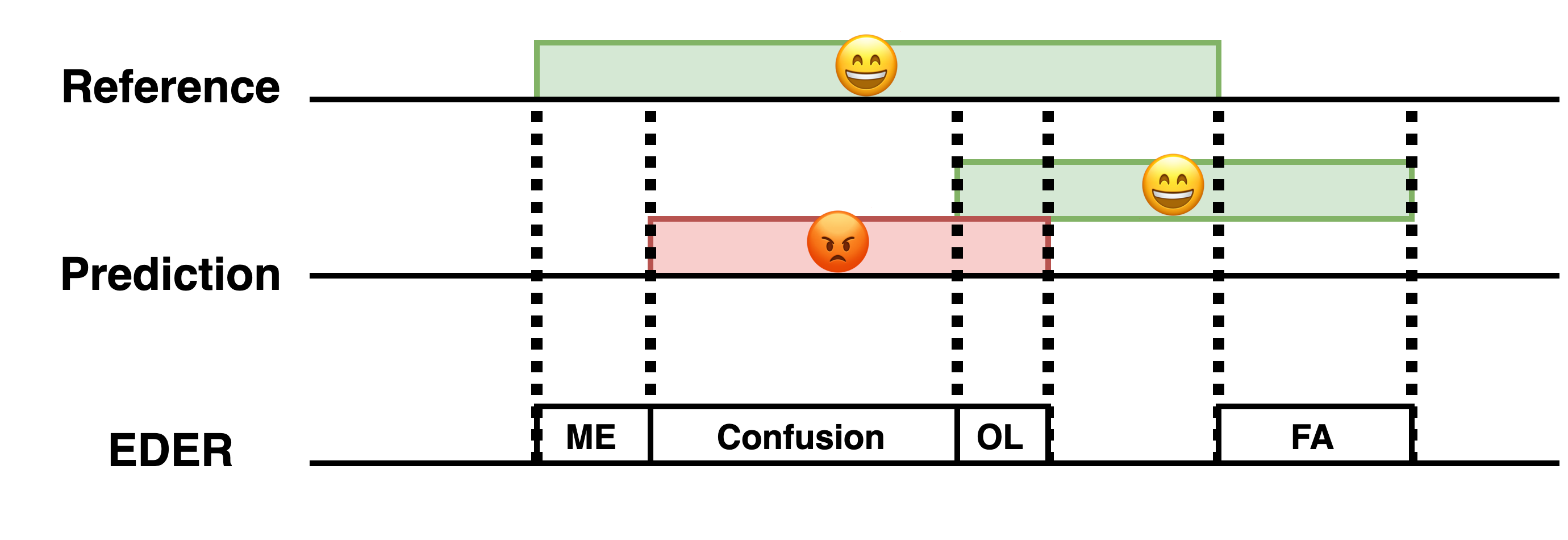}
  \caption{The four components of the Emotion Diarization Error Rate (EDER).}
  \label{fig:eder}
\end{figure}

The Emotion Diarization Error Rate (EDER) is specially designed to address the constraint of frame-level methods, which typically rely on utterance-level metrics for evaluation.
Inspired by the Diarization Error Rate (DER) used for Speaker Diarization, we define EDER as follows:
\begin{equation}
  EDER = \frac{FA + ME + CF + OL}{Utterance\ Duration}
\end{equation}
As shown in Figure \ref{fig:eder}, the EDER consists of the following components:
\begin{itemize}
    \item False Alarm (FA): The duration of non-emotional segments that are incorrectly predicted as emotional.
    \item Missed Emotion (ME): The duration of emotional segments that are incorrectly predicted as non-emotional.
    \item Confusion (CF): The duration of emotional segments that are incorrectly assigned to another (other) emotion(s).
    \item Overlap (OL): The duration of non-overlapped emotional segments that are predicted to contain other overlapped emotions apart from the correct one.
\end{itemize}

Establishing a standard benchmark can be challenging due to the use of different frame lengths in different frame-level solutions. However, by comparing the endpoints of the predicted emotion intervals and the actual emotion intervals, EDER takes the impact of frame lengths into consideration. Commonly, given the equivalent recognition capabilities of the models, the adoption of a smaller frame size provides finer time resolution and can subsequently result in lower EDER. The EDER metric also serves as a motivating factor for frame-level methods to strive for increased frame-wise accuracy as well as finer granularity.

%The proposed EDER metric is designed to precisely evaluate the temporal alignment between predicted emotion intervals and the ground truth emotion intervals. An implementation of EDER can also be found in our provided GitHub repository.

\section{Zaion Emotion Dataset}
\label{sec:Zaion Emotion Dataset}

One of the major challenges in studying fine-grained speech emotion recognition is the limited availability of datasets. To meet this lack, we built and released the Zaion Emotion Dataset (ZED), which includes discrete emotion labels and boundaries of the emotional segments within each utterance.

%The lack of available datasets is an important obstacle to the studies of fine-grained speech emotion recognition. Thus, we created the Zaion Emotion Dataset (ZED), which provides discrete emotion labels and the emotion boundaries at word-level for an utterance.

\subsection{Data Collection}
We first collected emotional YouTube videos that were licensed under the Creative Commons license. These videos were then converted from their original formats to single-channel 16 kHz WAV files.
The selected videos encompass a diverse range of emotional and interactive scenes, such as comedy shows, victim interviews, sports commentary, and more.
In the selected scenes, the speakers were fully aware of the environment and the context of the conversation.
The selected videos contain non-acted emotions recorded in real-life situations. This provides a unique set of challenges that are difficult to replicate in acted and laboratory conditions, offering a significant advantage in terms of dataset realism and relevance for industrial applications. Through the release of the ZED dataset, we aim to support the community in the evaluation and deployment of speech emotion recognition technologies in various real-life use cases.

%Emotional Youtube videos under the common-creative license were acquired and 
%. %The selection of videos was based on a wide range of emotional and interactive scenes such as comedy shows, victim interviews, sports comments, etc. In these selected scenes, the speakers were fully aware of the environment and the context of the conversation. The incorporation of non-acted/in-the-wild emotions makes the dataset more realistic and more relevant for industrial applications.

\subsection{Dataset Design}

We included three most frequently identified basic emotions \cite{ekman1971}: \textit{happy}, \textit{sad}, \textit{angry}. We also added \textit{neutral} to represent the non-emotional states that commonly exist in spoken utterances. 
% It should be noted that non-basic emotions  such as annoyance, pensiveness, and interest \cite{ekman} were all treated as \textit{neutral} in the ZED dataset.
%Neutral speech is perceived to be of limited value in both research and industry because it lacks emotional information that people are interested in. Hence, in the ZED dataset, \textit{neutral} is considered a non-emotional status that widely exists in spoken utterances. 
In contrast to neutrality, the other three emotions are considered as speech events. Detecting them and their boundaries is the goal of the proposed Speech Emotion Diarization task.
This task poses a significant challenge as the boundaries between different emotions are sometimes fuzzy even for human annotators.
%This task is very challenging because the boundaries of emotions are often fuzzy even for humans. I
To alleviate this difficulty, we limited our focus to the utterances that contained only one emotional event.
Figure \ref{fig:4mod} illustrates four possible transitional sequences for an utterance that exhibits only one emotional event \cite{emotion-ctc}. Each utterance of the designed ZED dataset corresponds to one of the four sequences.
%Figure \ref{fig:4mod} shows four valid transitions of emotional and non-emotional states available in the ZED dataset. As outlined before, only one non-neutral emotion is present \cite{emotion-ctc} in each recording. 

\subsection{Annotation Protocol}
\begin{figure}[t!]
  \centering
  \includegraphics[width=\linewidth]{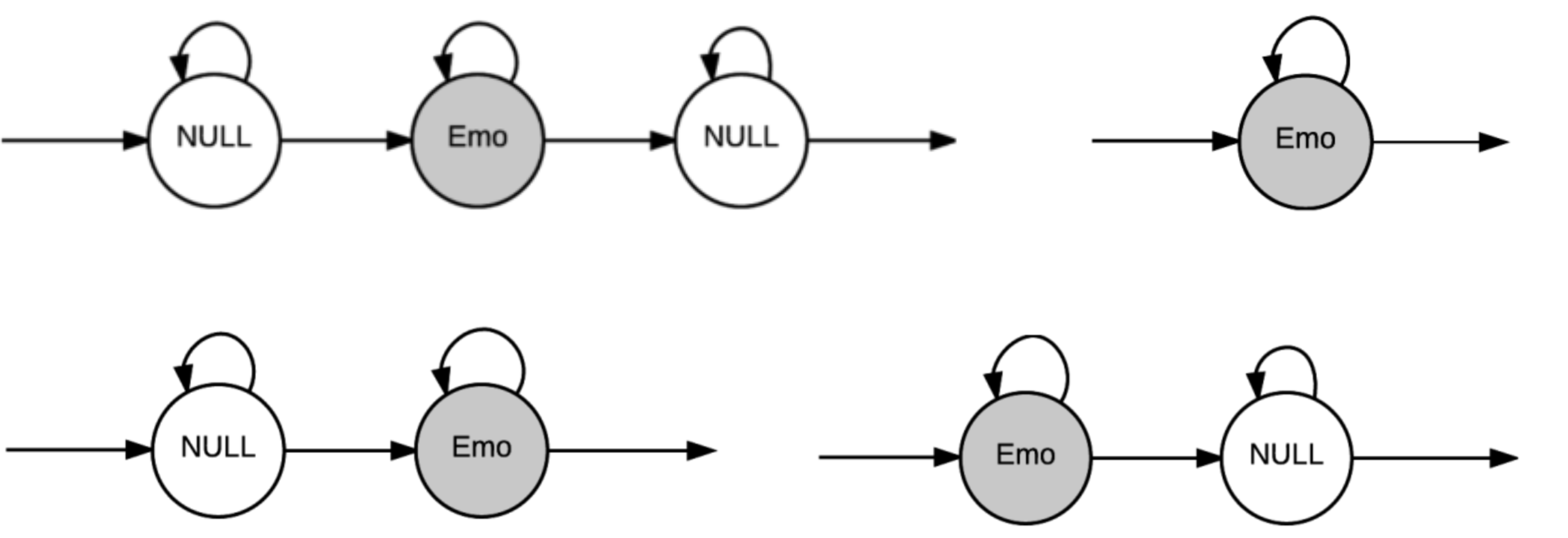}
  \caption{The four valid transitional sequences of an utterance where a single emotional event occurs \cite{emotion-ctc}. In the ZED dataset, the "Emo" label represents only one of the 3 emotions:  happiness, sadness, or anger. The "Null" label represents neutral states.}
  \label{fig:4mod}
\end{figure}

The annotation of the ZED dataset was conducted in a multi-level strategy, involving three main sessions that progressed from coarse to fine granularity:
\begin{itemize}
    \item utterance-level coarse-grained pre-selection,
    \item sub-utterance-level segmentation/annotation,
    \item frame-level boundary annotation.
\end{itemize}
First, the emotional speech recordings were segmented into utterances with a duration between 10 and 20 seconds. 
Three human experts worked on the utterance-level pre-selection by assigning a discrete emotion label to each utterance. Only those utterances annotated as \textit{happy}/\textit{sad}/\textit{angry} were passed to the second session. 
In the second session, five professional annotators with linguistic backgrounds performed the sub-utterance-level annotation. The utterances were first manually segmented into sub-utterances (if any), and then a discrete emotion label was assigned to each sub-utterance by the annotators. The annotations from the first session were verified during this process.
%After that, five professional annotators with linguistic background work on the sub-utterance/word level annotation. The utterances are first segmented into sub-utterances (if any), and then a discrete emotion label is assigned to each sub-sentence. During this process, the utterance-level annotations derived in the first session awere verified. 
Finally, a frame-level annotation was carried out to determine the durations of each emotional segment by the five annotators mentioned above. When the emotional boundaries needed to be determined, the annotators were asked to specify a precise timestamp within a frame as an endpoint of the emotion segment. For each session, the assignment of the emotional labels was based on the consensus derived from the annotators' subjective evaluations, a voting system was integrated to deal with the disagreements where the most voted category was selected as the final annotation. Only the utterances or sub-utterances that conform to the four sequences shown in Figure \ref{fig:4mod} were kept in the ZED dataset. Moreover, to reduce the negative impact of in-utterance silences on the quality of annotation, any utterance with silences longer than 0.2 seconds was excluded.
%Finally, a word-level annotation was conducted to detect the boundaries of each emotion segment. Only the utterances or sub-utterances that conform to the modalities shown in Figure \ref{fig:4mod} were kept. Also, to reduce the negative impact of in-sentence silences on the quality of annotation, we only kept sentences with silences shorter than 0.2 seconds.

% 问题：对于一个单词level来说，怎么判断是否属于某种情感; 

% The selection of the golden samples is based on the annotators' consensus.

\subsection{Dataset Statistics}

Table \ref{tab:stats} presents the overall statistics for the ZED dataset. It contains 180 utterances ranging from 1 to 15 seconds in duration and covers 73 speakers of different ages and genders. The proportion of the utterances corresponding to the three emotion categories is relatively balanced. Since there can be variations in the length of annotated emotion intervals across different annotators, we accessed the level of agreement by calculating the Fleiss' Kappa over short frames of 0.01 seconds. The obtained frame-wise Fleiss' Kappa is 0.81, which indicates a strong agreement among the five annotators. 
Additionally, the transcript of the speech content is also provided for each utterance.

% Since the Speech Emotion Diarization task requires the annotators to annotate both emotion categories and boundaries, there can be variations in the length of annotated emotion intervals across different annotators. In order to assess the level of agreement over emotional intervals of different lengths, we computed Fleiss' Kappa in a frame-wise setting. We assigned the annotated emotional categories to frames of 0.01 seconds and then calculated the Fleiss' Kappa over these short frames among the five annotators. The obtained frame-wise Fleiss' Kappa is 0.81, which indicates a strong agreement among the annotators.

\begin{table}[h!]
  \centering
  \caption{A summary of the overall statistics of the ZED dataset.
}
  \label{tab:stats}
  \centering
  \begin{tabular}{ll}
    \toprule
    % \multicolumn{1}{c}{\textbf{Model}} &   \multicolumn{1}{c}{\textbf{IC (ACC\%)}}\\
    % \midrule
    Language          & English \\
    % transcriptions          & Yes \\
    Number of utterances          & 180 \\
    Total duration     & 17 minutes  \\
    Number of speakers    & 73  \\
    Age of speakers    & 20 to 70 \\
    Gender of speakers    & 52\% female \\
    \bottomrule
    \% Happy & 34\% \\
    \% Sad & 37\% \\
    \% Angry & 29\% \\
    % emotions    & 34\%\textit{happy}-37\%\textit{sad}-29\%\textit{angry}\\    
    \bottomrule
  \end{tabular}
  
\end{table}
% 解释一下数据, with transcription 

% \begin{figure}[h]
%   \centering
%   \includegraphics[width=\linewidth]{emotions.png}
%   \caption{Emotions.}
%   \label{fig:emotions}
% \end{figure}

 %For now, ZED is a fairly small dataset, making it hard to use it as a training set. Nevertheless, we believe that thanks to the high-quality annotation conducted on in-the-wild emotional recordings, the ZED can serve as an excellent test set, as it can precisely assess the performance of an emotion diarization system in a very realistic and challenging task. 

 The ZED dataset is currently a small dataset, making it not ideal for training purposes. However, the inclusion of real-life emotions as well as the high-quality, fine-grained annotations make it an exceptional dataset for evaluation.
 % It can provide an accurate assessment of the system's capabilities .

\section{Experiments}
\label{sec:Experiments}

\subsection{Training Set}

% delete never, as zed is not enough for training, as far as we know

Since the current ZED dataset is not ideal for training, we constructed the training set using simulated data. We collected five English emotion datasets that are commonly used and annotated at utterance-level: IEMOCAP \cite{IEMOCAP}, RAVDESS \cite{RAVDESS}, Emov-DB \cite{Emov-DB}, ESD \cite{ESD}, and JL-CORPUS \cite{JL-CORPUS}, and we only selected the recordings that were annotated as \textit{happy}, \textit{sad}, \textit{angry}, and \textit{neutral}. The selected recordings were resampled to 16 kHz and were pre-processed by a Voice Activity Detection (VAD) module to confirm that no audio files contained silences lasting more than 0.2 seconds. For each processed recording, we assumed that the utterance-level emotional label could be perfectly attributed to the entirety of the utterance. Then, to obtain the training set, we randomly concatenated the processed recordings from the same speaker into longer utterances according to the four sequences shown in Figure \ref{fig:4mod}, assigning an equal probability of 0.25 to each. At last, a total of over 21 hours of simulated data were collected accounting for both the training and validation sets.

\subsection{Experimental Setup}

Large models (with about 317 million parameters) of pre-trained Wav2vec 2.0/HuBERT/WavLM were employed as emotional encoders in our experiments. These self-supervised models share the same architecture of the CNN feature encoder, for audios sampled at 16 kHz, this leads to a receptive field of 25ms and a stride of about 20ms between adjacent frames \cite{wav2vec2}. Despite the small receptive field, the transformer blocks inside the emotional encoder are able to effectively capture long-range dependencies and contextual emotional information thanks to the self-attention mechanism. 
Each encoded speech unit from the self-supervised model is then classified into one of the four emotion classes through a single fully connected layer.
For each utterance of the training set and the ZED dataset, the ground truth frame-wise labels were generated using the same 20ms stride. If two or more adjacent emotions occur within a frame, the frame will be assigned the emotion with the longer-lasting presence.
% It is noteworthy that under the condition of small frame lengths, the frame-wise classification error rate is close to the Emotion Diarization Error Rate.

The training process followed the methodology outlined in \cite{benchmark}. The CNN-based feature encoder was frozen while only the transformer blocks were fine-tuned on the downstream frame-wise classification task.
Two different schedulers were applied to adjust the learning rate of the self-supervised encoder and the learning rate of the downstream classifier.
For both schedulers, an Adam Optimizer was utilized and the learning rates were linearly annealed according to the performance of the validation stage. 
The self-supervised fine-tuning learning rate and the downstream learning rate were initialized to $10^{-5}$ and $10^{-4}$. 
The Negative Log-Likelihood (NLL) loss was calculated over the frame-wise classification and was then optimized.

\subsection{Results}

% post-processed results are to be added.

\begin{table}[t!]
  \centering
  \caption{Average Emotion Diarization Error Rate (EDER\%) and standard deviation (\%) of the proposed baseline models on Zaion Emotion Dataset computed over 5 different seeds.
}
  \label{tab:result}
  \centering
  \begin{tabular}{ll}
    \toprule
    \multicolumn{1}{c}{\textbf{Emotional Encoder}} &  \multicolumn{1}{c}{\textbf{EDER\%}}  \\
    \midrule
    Wav2vec2.0-large           & 36.2 $\pm$ 1.14  \\
    HuBERT-large               & 34.5 $\pm$ 0.81  \\
    WavLM-large                & \textbf{30.2} $\pm$ 1.60 \\
    \bottomrule
  \end{tabular}
\end{table}

The performance of the proposed baseline models is shown in Table \ref{tab:result}. The baselines were trained with a 90/10 train-validation split of the constructed training set and were evaluated on the proposed ZED dataset. As the ZED dataset remains fixed and separate from the training/validation set, the evaluation was conducted under a speaker-independent setup. The reported EDER values were averaged across 5 different train-validation splitting seeds. In addition, the standard deviation is also reported to indicate the range of variation.

It can be inferred from the obtained results that all three models achieve good performance with low EDERs. Specifically, the WavLM-large model outperforms the Wav2vec2.0-large and HuBERT-large models and reaches an EDER of 30.2\%, suggesting its better suitability for speech emotion recognition tasks than the other two models.
% Interestingly, the superiority of WavLM for speech emotion recognition has also been observed in other studies, including the SUPERB challenge \cite{}. %Therefore, these findings suggest that WavLM could be a powerful tool for a wide range of speech processing tasks beyond speech emotion recognition.

%The performance of the proposed baseline models is shown in Table \ref{tab:result}. The reported EDER is averaged over 5 different seeds with an 80/20 train-validation split. The corresponding standard deviation is also reported. The results reveal that all three models reach a relatively good EDER performance. WavLM-large outperforms Wav2vec2.0-large and HuBERT-large, achieving a EDER of 30.2\%. The superiority of WavLM for speech emotion recognition (and for many other speech processing tasks) has also been observed in the context of the SUPERB challenge. 

To further showcase the dynamic modeling capability of the model, we report the performance of the WavLM-large model on the prediction of the four emotion transitions in Table \ref{tab:example}. For each utterance, if the predicted emotions remain constant across adjacent frames, we fuse these predictions to obtain a shorter emotion sequence. We then compare the obtained sequences with the ground truth sequence shown in Figure \ref{fig:4mod}. This aims to determine whether the model is able to correctly identify the approximate location of the emotion, i.e., the entirety, beginning, middle, or end of the utterance, and to further provide insights into the model's general capability.
The results demonstrate that the model correctly identifies the emotion transition in 42\% of the predictions. 
In particular, we observe a decreasing accuracy as the number of emotion transitions increases in the utterance. Especially when the emotional event appears in the middle of the utterance, the model produces poor predictions, which indicates a clear increase in the difficulty of the task. 
Despite the inherent challenges present in the ZED dataset and the use of simulated training data, the results achieved by our model were satisfactory. Our finding provides evidence of the model's capability on distinguishing emotions at a fine temporal granularity and also validates the feasibility of the proposed Speech Emotion Diarization task on the ZED dataset.

%Table \ref{tab:example} shows the performance of the WavLM-large model for the four emotion modalities. For each utterance, if the predicted emotion remains constant across adjacent frames, we fuse the predictions of these frames to get a shorter emotion sequence. We then compare the obtained sequence with the ground truth sequence (one of the four modalities in Figure \ref{fig:4mod}). The goal is to assess whether the model is able to determine the approximate location of the emotion (at the beginning/in the middle/at the end of the utterance), and to further provide insights into the model's general capability. The results show that in 42\% of the predictions, the emotion modality can be correctly identified. Even when using simulated data for traning, the obtained results are surprisingly, especially considering that the test signals are recorded in real-life conditions.

%which also provide evidence of the capability of the model and validate the feasibility of the Speech Emotion Diarization task. 

\begin{table}[t!]
  \centering
  \caption{
  Performance of the WavLM-large model on emotion transition prediction. For each emotion transition in the Zaion Emotion Dataset, we present the total number of utterances  (TN), the number of utterances with the correct emotion transition predicted (CN), and the accuracy (ACC) which is equal to $CN/TN$. 
  }
  \label{tab:example}
  \centering
  \begin{tabular}{llll}
    \toprule
    \multicolumn{1}{l}{\textbf{Transition}} &   \multicolumn{1}{l}{\textbf{TN}} &
    \multicolumn{1}{l}{\textbf{CN}}  &     
    \multicolumn{1}{l}{\textbf{ACC}}\\
    \midrule
    \textit{happy}          & 5 & 5 & 100\% \\
    \textit{null-happy}          & 31 & 14 & 45.2\% \\
    \textit{happy-null}     & 14 & 8  & 57.1\% \\
    \textit{null-happy-null}     & 12 & 3  & 25.0\% \\
    \midrule
    \textit{sad}          & 17 & 13 & 76.5\% \\
    \textit{null-sad}          & 33 & 7 & 21.2\% \\
    \textit{sad-null}     & 10 & 3  & 30.0\% \\
    \textit{null-sad-null}     & 6 & 0  & 0.0\% \\
    \midrule
    \textit{angry}          & 8 & 7 & 87.5\% \\
    \textit{null-angry}          & 30 & 15 & 50.0\% \\
    \textit{angry-null}     & 11 & 1  & 9.1\% \\
    \textit{null-angry-null}     & 3 & 0  & 0.0\% \\
    \midrule
    \textbf{Total} & 180 & 76 & 42.2\%  \\
    \bottomrule
  \end{tabular}
  
\end{table}

\section{Conclusions}
\label{sec:Conclusions}

In this work, we investigated fine-grained speech emotion recognition and proposed a Speech Emotion Diarization task that aims to identify the emotions along with their temporal boundaries. %The task was evaluated with the proposed Emotion Diarization Error Rate metric. 
We also introduced the Zaion Emotion Dataset, a freely available fine-grained emotion dataset containing real-life emotions. We provided performant baseline models, with the WavLM-large model achieving the lowest EDER of 30.2\%. Both the code and the pre-trained models have been open-sourced. Through our work, we hope to further reveal the fine-grained nature of speech emotions and to establish a common benchmark for fine-grained speech emotion recognition approaches through a diarization objective.

% In this study, our main focus was on fine-grained speech emotion recognition. To achieve this, we proposed a new task called Speech Emotion Diarization that aims to identify the emotions along with their temporal boundaries. We evaluated the performance using our proposed metric, the Emotion Diarization Error Rate.
% As part of this study, we also introduced the Zaion Emotion Dataset. This is a freely available dataset of fine-grained emotions, containing non-acted emotions recorded in real-life conditions. We also released our performant baseline models and provided the training recipes and pre-trained models in an open-source format. Our best model is based on 
% WavLM-large and achieves an EDER of 30.2\%.

% This work aims to establish a common benchmark that will encourage further research into fine-grained speech emotion recognition.
In our future work, we plan to expand the dataset to a larger scale and to include more languages. We also plan to explore the integration of front-end and post-processing techniques, as well as develop novel baseline models with lower EDER. 

\section{Acknowledgements}
We would like to thank Björn Schuller, Alaa Nfissi, and colleagues from Zaion Lab/Zaion Data Team for the helpful discussions.

\bibliographystyle{IEEEbib}
\bibliography{strings,refs}

\begin{thebibliography}{10}

\bibitem{survey}
Hatice Gunes, Bj{\"o}rn Schuller, Maja Pantic, and Roddy Cowie,
\newblock ``Emotion representation, analysis and synthesis in continuous space: A survey,''
\newblock in {\em 2011 IEEE International Conference on Automatic Face \& Gesture Recognition (FG)}. IEEE, 2011, pp. 827--834.

\bibitem{review1}
Ruhul~Amin Khalil, Edward Jones, Mohammad~Inayatullah Babar, Tariqullah Jan, Mohammad~Haseeb Zafar, and Thamer Alhussain,
\newblock ``Speech emotion recognition using deep learning techniques: A review,''
\newblock {\em IEEE Access}, vol. 7, pp. 117327--117345, 2019.

\bibitem{review2}
Taiba~Majid Wani, Teddy~Surya Gunawan, Syed Asif~Ahmad Qadri, Mira Kartiwi, and Eliathamby Ambikairajah,
\newblock ``A comprehensive review of speech emotion recognition systems,''
\newblock {\em IEEE Access}, vol. 9, pp. 47795--47814, 2021.

\bibitem{review3}
Mehmet~Berkehan Ak{\c{c}}ay and Kaya O{\u{g}}uz,
\newblock ``Speech emotion recognition: Emotional models, databases, features, preprocessing methods, supporting modalities, and classifiers,''
\newblock {\em Speech Communication}, vol. 116, pp. 56--76, 2020.

\bibitem{capsule}
Xixin Wu, Songxiang Liu, Yuewen Cao, Xu~Li, Jianwei Yu, Dongyang Dai, Xi~Ma, Shoukang Hu, Zhiyong Wu, Xunying Liu, et~al.,
\newblock ``Speech emotion recognition using capsule networks,''
\newblock in {\em ICASSP 2019-2019 IEEE International Conference on Acoustics, Speech and Signal Processing (ICASSP)}. IEEE, 2019, pp. 6695--6699.

\bibitem{benchmark}
Yingzhi Wang, Abdelmoumene Boumadane, and Abdelwahab Heba,
\newblock ``A fine-tuned wav2vec 2.0/hubert benchmark for speech emotion recognition, speaker verification and spoken language understanding,''
\newblock {\em arXiv preprint arXiv:2111.02735}, 2021.

\bibitem{spk-norm}
Itai Gat, Hagai Aronowitz, Weizhong Zhu, Edmilson Morais, and Ron Hoory,
\newblock ``Speaker normalization for self-supervised speech emotion recognition,''
\newblock in {\em ICASSP 2022-2022 IEEE International Conference on Acoustics, Speech and Signal Processing (ICASSP)}. IEEE, 2022, pp. 7342--7346.

\bibitem{dim1}
Didier Grandjean, David Sander, and Klaus~R Scherer,
\newblock ``Conscious emotional experience emerges as a function of multilevel, appraisal-driven response synchronization,''
\newblock {\em Consciousness and cognition}, vol. 17, no. 2, pp. 484--495, 2008.

\bibitem{dim2}
James~A Russell,
\newblock ``A circumplex model of affect.,''
\newblock {\em Journal of personality and social psychology}, vol. 39, no. 6, pp. 1161, 1980.

\bibitem{pad}
Albert Mehrabian,
\newblock ``Pleasure-arousal-dominance: A general framework for describing and measuring individual differences in temperament,''
\newblock {\em Current Psychology}, vol. 14, pp. 261--292, 1996.

\bibitem{dimensional1}
Martin W{\"o}llmer, Bj{\"o}rn Schuller, Florian Eyben, and Gerhard Rigoll,
\newblock ``Combining long short-term memory and dynamic bayesian networks for incremental emotion-sensitive artificial listening,''
\newblock {\em IEEE Journal of selected topics in signal processing}, vol. 4, no. 5, pp. 867--881, 2010.

\bibitem{dimensional2}
Martin W{\"o}llmer, Florian Eyben, Stephan Reiter, Bj{\"o}rn Schuller, Cate Cox, Ellen Douglas-Cowie, and Roddy Cowie,
\newblock ``Abandoning emotion classes-towards continuous emotion recognition with modelling of long-range dependencies,''
\newblock 2008.

\bibitem{SEWA}
Jean Kossaifi, Robert Walecki, Yannis Panagakis, Jie Shen, Maximilian Schmitt, Fabien Ringeval, Jing Han, Vedhas Pandit, Antoine Toisoul, Bj{\"o}rn Schuller, et~al.,
\newblock ``Sewa db: A rich database for audio-visual emotion and sentiment research in the wild,''
\newblock {\em IEEE transactions on pattern analysis and machine intelligence}, vol. 43, no. 3, pp. 1022--1040, 2019.

\bibitem{RECOLA}
Fabien Ringeval, Andreas Sonderegger, Juergen Sauer, and Denis Lalanne,
\newblock ``Introducing the recola multimodal corpus of remote collaborative and affective interactions,''
\newblock in {\em 2013 10th IEEE international conference and workshops on automatic face and gesture recognition (FG)}. IEEE, 2013, pp. 1--8.

\bibitem{SEMAINE}
Gary McKeown, Michel Valstar, Roddy Cowie, Maja Pantic, and Marc Schroder,
\newblock ``The semaine database: Annotated multimodal records of emotionally colored conversations between a person and a limited agent,''
\newblock {\em IEEE transactions on affective computing}, vol. 3, no. 1, pp. 5--17, 2011.

\bibitem{categorical}
Paul Ekman and Wallace~V Friesen,
\newblock ``Unmasking the face: A guide to recognizing emotions from facial clues.,''
\newblock 1975.

\bibitem{emotion-ctc}
Vladimir Chernykh and Pavel Prikhodko,
\newblock ``Emotion recognition from speech with recurrent neural networks,''
\newblock {\em arXiv preprint arXiv:1701.08071}, 2017.

\bibitem{video}
Tianyi Zhang,
\newblock ``On fine-grained temporal emotion recognition in video: How to trade off recognition accuracy with annotation complexity?,''
\newblock 2022.

\bibitem{SpeechBrain}
Mirco Ravanelli, Titouan Parcollet, Peter Plantinga, Aku Rouhe, Samuele Cornell, Loren Lugosch, Cem Subakan, Nauman Dawalatabad, Abdelwahab Heba, Jianyuan Zhong, et~al.,
\newblock ``Speechbrain: A general-purpose speech toolkit,''
\newblock {\em arXiv preprint arXiv:2106.04624}, 2021.

\bibitem{framevsturn}
Bogdan Vlasenko, Bj{\"o}rn Schuller, Andreas Wendemuth, and Gerhard Rigoll,
\newblock ``Frame vs. turn-level: emotion recognition from speech considering static and dynamic processing,''
\newblock in {\em Affective Computing and Intelligent Interaction: Second International Conference, ACII 2007 Lisbon, Portugal, September 12-14, 2007 Proceedings 2}. Springer, 2007, pp. 139--147.

\bibitem{extreme}
Kun Han, Dong Yu, and Ivan Tashev,
\newblock ``Speech emotion recognition using deep neural network and extreme learning machine,''
\newblock in {\em Interspeech 2014}, 2014.

\bibitem{realtime}
Haytham~M Fayek, Margaret Lech, and Lawrence Cavedon,
\newblock ``Towards real-time speech emotion recognition using deep neural networks,''
\newblock in {\em 2015 9th international conference on signal processing and communication systems (ICSPCS)}. IEEE, 2015, pp. 1--5.

\bibitem{local-attention}
Seyedmahdad Mirsamadi, Emad Barsoum, and Cha Zhang,
\newblock ``Automatic speech emotion recognition using recurrent neural networks with local attention,''
\newblock in {\em 2017 IEEE International Conference on Acoustics, Speech and Signal Processing (ICASSP)}, 2017, pp. 2227--2231.

\bibitem{attention-pooling}
Pengcheng Li, Yan Song, Ian~Vince McLoughlin, Wu~Guo, and Li-Rong Dai,
\newblock ``An attention pooling based representation learning method for speech emotion recognition,''
\newblock 2018.

\bibitem{evaluating}
Haytham~M Fayek, Margaret Lech, and Lawrence Cavedon,
\newblock ``Evaluating deep learning architectures for speech emotion recognition,''
\newblock {\em Neural Networks}, vol. 92, pp. 60--68, 2017.

\bibitem{IEMOCAP}
Carlos Busso, Murtaza Bulut, Chi-Chun Lee, Abe Kazemzadeh, Emily Mower, Samuel Kim, Jeannette~N Chang, Sungbok Lee, and Shrikanth~S Narayanan,
\newblock ``Iemocap: Interactive emotional dyadic motion capture database,''
\newblock {\em Language resources and evaluation}, vol. 42, no. 4, pp. 335--359, 2008.

\bibitem{RAVDESS}
Steven~R Livingstone and Frank~A Russo,
\newblock ``The ryerson audio-visual database of emotional speech and song (ravdess): A dynamic, multimodal set of facial and vocal expressions in north american english,''
\newblock {\em PloS one}, vol. 13, no. 5, pp. e0196391, 2018.

\bibitem{EMO-DB}
Felix Burkhardt, Astrid Paeschke, Miriam Rolfes, Walter~F Sendlmeier, Benjamin Weiss, et~al.,
\newblock ``A database of german emotional speech.,''
\newblock in {\em Interspeech}, 2005, vol.~5, pp. 1517--1520.

\bibitem{attention}
Ashish Vaswani, Noam Shazeer, Niki Parmar, Jakob Uszkoreit, Llion Jones, Aidan~N Gomez, {\L}ukasz Kaiser, and Illia Polosukhin,
\newblock ``Attention is all you need,''
\newblock {\em Advances in neural information processing systems}, vol. 30, 2017.

\bibitem{wav2vec2}
Alexei Baevski, Yuhao Zhou, Abdelrahman Mohamed, and Michael Auli,
\newblock ``wav2vec 2.0: {A} framework for self-supervised learning of speech representations,''
\newblock in {\em NeurIPS}, 2020.

\bibitem{HuBERT}
Wei-Ning Hsu, Benjamin Bolte, Yao-Hung~Hubert Tsai, Kushal Lakhotia, Ruslan Salakhutdinov, and Abdelrahman Mohamed,
\newblock ``Hubert: Self-supervised speech representation learning by masked prediction of hidden units,''
\newblock {\em IEEE/ACM Transactions on Audio, Speech, and Language Processing}, vol. 29, pp. 3451--3460, 2021.

\bibitem{WavLM}
Sanyuan Chen, Chengyi Wang, Zhengyang Chen, Yu~Wu, Shujie Liu, Zhuo Chen, Jinyu Li, Naoyuki Kanda, Takuya Yoshioka, Xiong Xiao, et~al.,
\newblock ``Wavlm: Large-scale self-supervised pre-training for full stack speech processing,''
\newblock {\em IEEE Journal of Selected Topics in Signal Processing}, vol. 16, no. 6, pp. 1505--1518, 2022.

\bibitem{wavlm-emo}
Kuo-Hsuan Hung, Szu-wei Fu, Huan-Hsin Tseng, Hsin-Tien Chiang, Yu~Tsao, and Chii-Wann Lin,
\newblock ``Boosting self-supervised embeddings for speech enhancement,''
\newblock {\em Interspeech 2022}, 2022.

\bibitem{ekman1971}
Paul Ekman,
\newblock ``Universals and cultural differences in facial expressions of emotion.,''
\newblock in {\em Nebraska symposium on motivation}. University of Nebraska Press, 1971.

\bibitem{Emov-DB}
Adaeze Adigwe, No{\'e} Tits, Kevin~El Haddad, Sarah Ostadabbas, and Thierry Dutoit,
\newblock ``The emotional voices database: Towards controlling the emotion dimension in voice generation systems,''
\newblock {\em arXiv preprint arXiv:1806.09514}, 2018.

\bibitem{ESD}
Kun Zhou, Berrak Sisman, Rui Liu, and Haizhou Li,
\newblock ``Seen and unseen emotional style transfer for voice conversion with a new emotional speech dataset,''
\newblock in {\em ICASSP 2021-2021 IEEE International Conference on Acoustics, Speech and Signal Processing (ICASSP)}. IEEE, 2021, pp. 920--924.

\bibitem{JL-CORPUS}
Jesin James, Li~Tian, and Catherine Watson,
\newblock ``An open source emotional speech corpus for human robot interaction applications,''
\newblock {\em Interspeech 2018}, 2018.

\end{thebibliography}

\end{document}